\documentclass[final]{cvpr}

\usepackage{times}
\usepackage{epsfig}
\usepackage{graphicx}
\usepackage{amsmath}
\usepackage{amssymb}
\usepackage{multirow}

\usepackage[pagebackref=true,breaklinks=true,colorlinks,bookmarks=false]{hyperref}

\def\cvprPaperID{8377} 
\def\confYear{CVPR 2024}

\begin{document}

\title{Glass Segmentation with Multi Scales and Primary Prediction Guiding}

\author{Zhiyu Xu\\
College of Information Science and Technology\\
Jinan University\\
{\tt\small xuzhiyu@stu2022.jnu.edu.cn}
\and
Qingliang Chen\thanks{Corresponding author} \\
Department of Computer Science\\
Jinan University\\
{\tt\small tpchen@jnu.edu.cn}
}

\maketitle

\begin{abstract}
Glass-like objects can be seen everywhere in our daily life which are very hard for existing methods to segment them. The properties of transparencies pose great challenges of detecting them from the chaotic background and the vague separation boundaries further impede the acquisition of their exact contours. Moving machines which ignore glasses have great risks of crashing into transparent barriers or difficulties in analysing objects reflected in the mirror, thus it is of substantial significance to accurately locate glass-like objects and completely figure out their contours. In this paper, inspired by the scale integration strategy and the refinement method, we proposed a brand-new network, named as \textbf{MGNet}, which consists of a Fine-Rescaling and Merging module (\textbf{FRM}) to improve the ability to extract spatially relationship and a Primary Prediction Guiding module (\textbf{PPG}) to better mine the leftover semantics from the fused features. Moreover, we supervise the model with a novel loss function with the uncertainty-aware loss to produce high-confidence segmentation maps. Unlike the existing glass segmentation models that must be trained on different settings with respect to varied datasets, our model are trained under consistent settings and has achieved superior performance on three popular public datasets. Code is available at   
\end{abstract}

\section{Introduction}
 Glass objects are known to be very hard for segmentation which possess two main characteristics: transparency and specular reflection. These two attributes endow glass with a substantial challenge in visual recognition, making it susceptible to blending with the background. Moreover, the presence of indistinct and gradient boundaries further complicate the detection and segmentation process. In practices, failure to detect glasses can potentially lead to collisions with their barriers for machines such as robots and autonomous vehicles, as well as hindering the tracking of objects. Consequently, new powerful methods and models for the detection and segmentation of glasses are badly needed.

 \begin{figure}[t]
 
    \centering
    \includegraphics[width=1\linewidth,height=6.0cm]{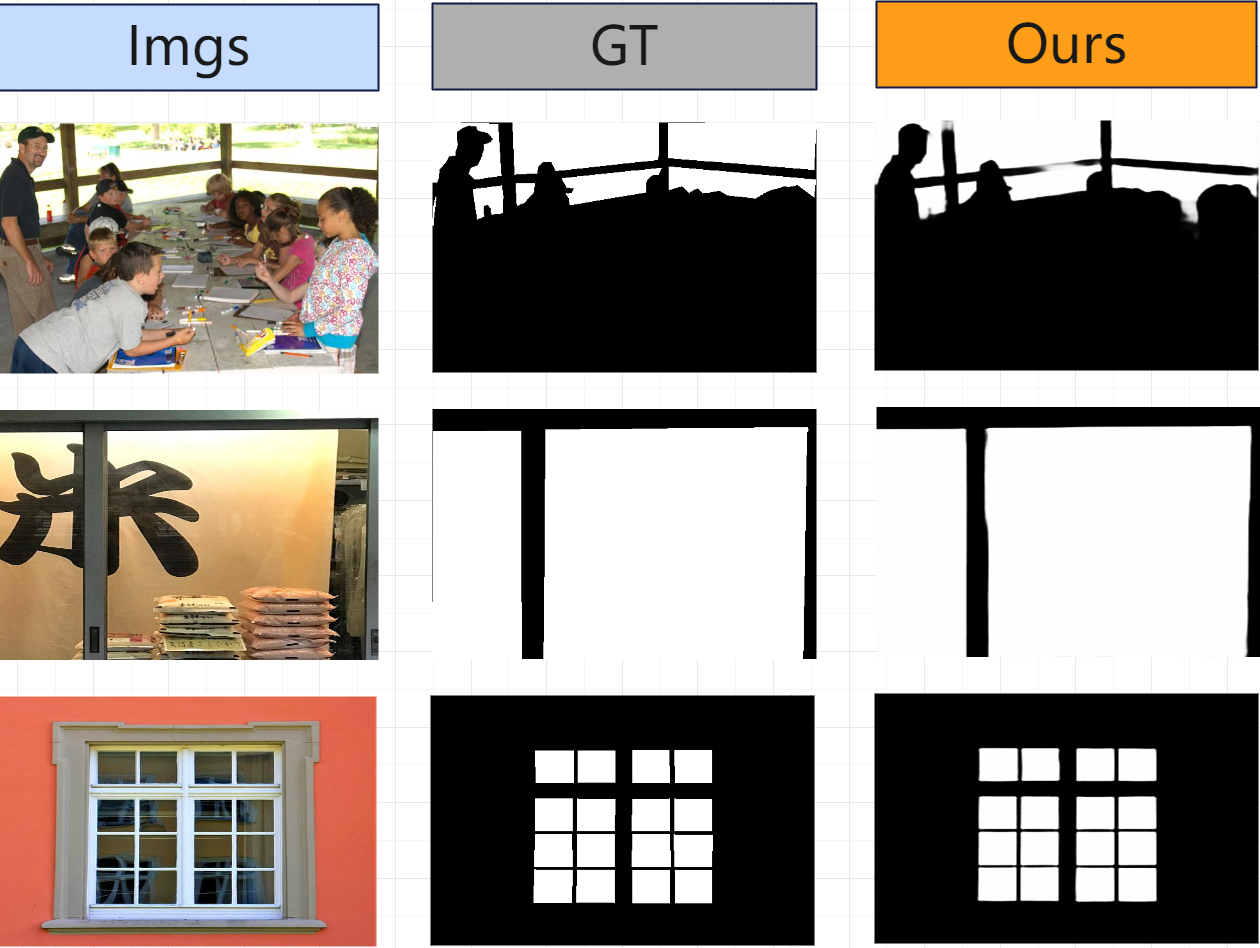}
    \caption{Visualization of our predicted results. Compared with the ground truths, our segmentation maps exhibit fine, complete, and strong discriminative properties, which is mainly contributed by our combination of multi-scales and refinement strategies. The predictions of those ambiguous regions also gain the help from UAL.}

    \label{fig:Figure 1}

\end{figure}

The existing segmentation methods for transparent materials mainly leverage boundary detection for assistance ~\cite{mei2020don,fan2023rfenet}. Furthermore, some methods have been proposed to detect and segment glass using new attributes such as Spectral Polarization ~\cite{mei2022glass}. However, it is evident that these methods introduce additional computational overhead, and these attributes are not universally present in every glass object. For instance, some mirrors have well-defined boundaries, while certain glass exhibits perplexing edges. Additionally, acquiring other attributes, such as spectral polarization information, for each object is not always convenient. For the glass segmentation task, we intend to rely solely on conventional RGB information, similar to the approach taken for other objects. This approach not only simplifies the data acquisition and processing but also facilitates a better analysis, thereby improving model applicability.

In this paper, we are dedicated to addressing the problems arising from object interference within glasses and the challenging segmentation of objects due to the special boundaries of glasses in the background. Drawing inspiration from widely studied and adopted multi-scale input strategies and iterative refinement modules, we propose a novel network architecture, referred to as MGNet, which includes a Fine-Rescaling and Merging module (FRM) and a Primary Prediction Guiding module (PPG). FRM makes slight adjustments to the original input image's dimensions, enhancing the spatial relationship among objects within the image and capturing more abstract representations. Moreover, appropriate adjustments sharpen the image's edges in one instance (with reduced size) and dull them in another instance (with an enlarged size). Through an attention-generating gate, each pixel in features extracted from the images are weighted, significantly enhancing semantic representations. And PPG aims to address the challenge of incomplete and low-confidence segmentation posed by the ambiguous boundaries between the glass-like objects and the background, mining the leftover semantics from the final features by guiding the segmentation process with the primary prediction. Furthermore, to enhance the effectiveness of eliminating the uncertainty regions (low-confidence regions), we utilize the  Uncertainty-Aware Loss (UAL) proposed by ~\cite{pang2022zoom} as an auxiliary loss function. 

Our contributions can be summarized as follows:
\begin{itemize}
\item We propose MGNet, a novel glass-like object segmentation model, which combines two empirically effective strategies of multi-scales and refinement.

\item We propose a Fine-Rescaling and Merging module (FRM) to improve the ability of our model to extract spatially relations and a Primary Prediction Guiding module (PPG) to better mining the leftover semantics from the fused features.

\item Our model has been implemented and achieved state-of-the-art performance on two popular benchmarks and superior performance on another instance segmentation dataset under the consistent train settings, presenting better transferability since existing models must adjust their settings respect to each benchmark to fit their methods for all datasets.
  
\end{itemize}

\section{Related Work}
\textbf{Glass-like Object Segmentation.}  Segmenting objects with a glass-like appearance presents a significantly greater challenge compared to commonly seen objects and this heightened difficulty arises primarily from the fact that the inner regions of glass objects often exhibit a perplexing similarity to their surrounding backgrounds. To address this issue, some methods ~\cite{mei2022glass} have turned to the utilization of additional multi-modal information, such as 4D light-field data, refractive flow maps, thermal imaging and spectral polarization. Regrettably, the acquisition of such multi-modal data is relatively expensive, thus limiting its broader applicability. Instead, recent contributions from researchers such as ~\cite{mei2020don,lin2021rich,xie2020segmenting,lin2020progressive}, have led to the creation of large-scale RGB image datasets specifically tailored to glass-like objects, promoting this research in the community. However, due to the unique characteristics of glass-like objects, conventional existing semantic segmentation methods ~\cite{chen2018encoder,zhao2017pyramid}, have failed to deliver promising results. Similarly, many cutting-edge approaches in salient object detection ~\cite{pang2020multi,pang2022zoom} have also struggled, as glass objects may not always exhibit salient features.
Additionally, ~\cite{xie2020segmenting,lin2021rich} have introduced methods for the segmentation of glass-like objects with the aid of boundary cues, leveraging the precise localization afforded by boundaries. Unlike the methods mentioned above, our approach only processes the raw RGB images with a novel network combining two strategies of multi-scales and refinement without any  boundary computation.  

\textbf{Scale Space Integration.} Scale-space theory typically aims to address natural variations in scales by achieving an optimal comprehension of image structure, providing an exceedingly effective and theoretically sound framework. This concept has found widespread applications in computer vision, including the utilization of image pyramids ~\cite{adelson1984pyramid} and feature pyramids ~\cite{lin2017feature}. Recent CNN-based methods for Camouflaged Object Detection (COD) ~\cite{pang2022zoom,fan2020camouflaged} and Salient Object Detection (SOD) ~\cite{pang2020multi,ji2022dmra,pang2020hierarchical} have explored strategies that combine inter-layer features to enhance feature representations. These approaches have demonstrated positive impacts on accurate object localization and segmentation. However, for the glass segmentation task, the existing approaches overlook the substantial effect of varying input scales on detecting glass-like objects and thus they are often confused by the complex spatial relationships in images between objects. In this paper, we apply the mixed-scale integration strategy to adjust the attention of our model to the scales it needs most with respect to each pixel to significantly improve the ability of our model to figure out the structure of input images. Furthermore, we design our network along with the fine-rescale and merging module and the hierarchical channel-down decoder which effectively extract and fuse the semantics from different scales.

\textbf{Iterative Refinement.} The iterative refinement aims to optimize the secondary structure and elaborate details through repeating
the final or nearly final process of feature extraction supervised by the results predicted in the last step. Recently,this strategy has been
widely utilized on few-shot learning~\cite{zhang2019canet} and camouflaged object detection as well~\cite{jia2022segment}. Jia et al. propose an attention-based iterative refinement module in ~\cite{jia2022segment}, which samples the target area iteratively to improve the ability of the model to detect the objects in small scales. Unlike the models mentioned above, our proposed iterative refinement module intends to make our model segment objects in a coarse way firstly and then guide the primary predicted results with the final features to fully mine the semantics left in the processed final features. Therefore our model is able to detect the severely glass-like objects and generates final prediction which is more structured and regular.     

\begin{figure*}[ht]
    \centering
    \includegraphics[width=1.0\linewidth,height=5cm]{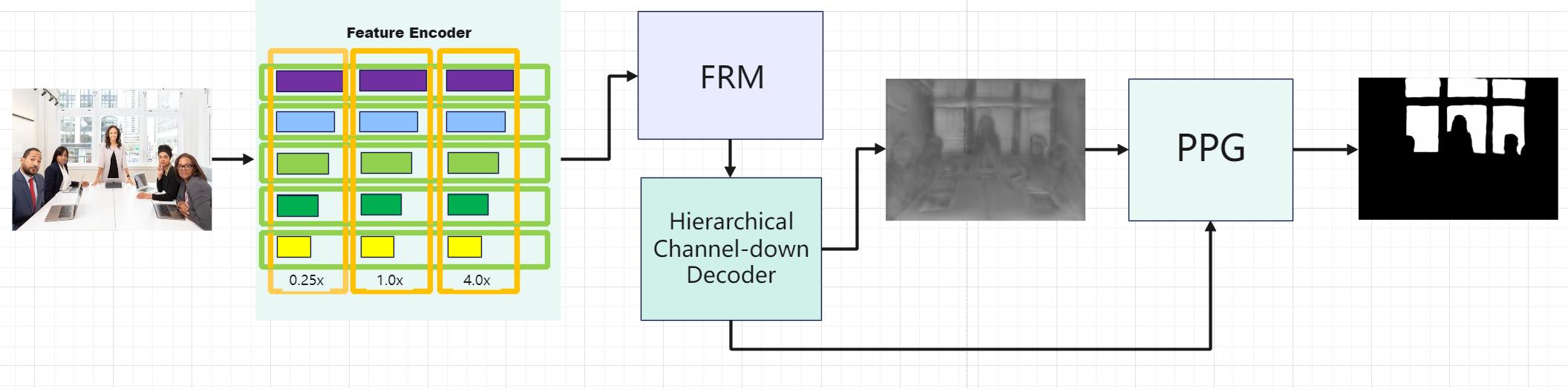}
    \caption{The overall framework of MGNet. The fine-rescaling and merging module (\textbf{FRM}) is adopted to integrate full-channel features of different levels extracted by a shared encoder to mine the critical clues from different scales. The hierarchical channel-down decoder (\textbf{HCDD}) further enhances the feature discrimination by constructing a multi-path structure inside the features while reducing the channel of features. Then, a coarse probability map is generated and sent to the primary prediction guiding module (\textbf{PPG}) to complete the coarse-to-fine process for mining leftover semantics and uncertainty elimination. After refinement, a final probability map of the camouflaged object on the input image can be obtained. The times of guiding can be adjusted according to practical needs.}
    \label{fig:Figure 2}
\end{figure*}

\section{ Methodology}
  In this section, we first elaborate on the overall architecture of the proposed MGNet, and then present the details of each module.

\subsection{Overall Architecture}

The overall architecture of the proposed MGNet is illustrated in Fig.~\ref{fig:Figure 2}. Inspired by the rescaling strategy and the rechecking strategy of human beings, we better and advance those strategies for getting performance promotion in glass-like object segmentation. By consolidating information from various scales, we can effectively uncover subtle but valuable cues in complex situations, ultimately aiding in the process of glass-like objects segmentation, thus we finely adjust the scales of input image (height and weight), i.e., 0.7x, 1.0x, 1.2x, which conform to our ideas of multi-scales input. The images of different scales share a feature encoder to extract different levels of features and the feature encoder we use in this paper is the pre-trained ResNeXt101. The features of different scales are organized based on their respective channels, and these groups are then directed to the Fine-Rescaling and Merging module (FRM), which is based on the attention-aware filtering mechanism to integrate features that contain rich scale-specific information. The module will generate attention to different scales which fits the picture being detected and significantly improve the model's ability to capture crucial and informative semantic cues for detecting challenging, glass-like objects. After that, the fused multi-level features are gradually integrated through Hierarchical Channel-Down moDule (HCDD) in a top-down manner to enhance the mixed-scale feature representation while their channel is reduced accordingly to be able to mix with others hierarchically. It further expands the receptive field ranges and integrates information from the semantic level to the local level. At the end of the integration process, a coarse logits map can be obtained. To eliminate the uncertainty in the coarse logits map and further mine the leftover information of structures and local details of the final fused features, we employ the primary prediction guiding module to process the feature generated by the last module (HCDD) with the primary logits map. After the refinement, the final segmentation map is generated by the subsequent logits map through a convolutional layer and it is supervised by the ground truth. The loss function is composed of the binary cross entropy (BCE) loss and an uncertainty-aware loss (UAL) to enable the model to eliminate the uncertain regions and produce a structured, regular and accurate prediction. 

\subsection{Feature Encoder}
We first extract deep features through a shared encoder and group them by their channels. For fully implement our idea of finely adjusting scales to promote spatially relationship input, 
\begin{figure}[ht]
    \centering
    \includegraphics[width=1.0\linewidth,height=3.5cm]{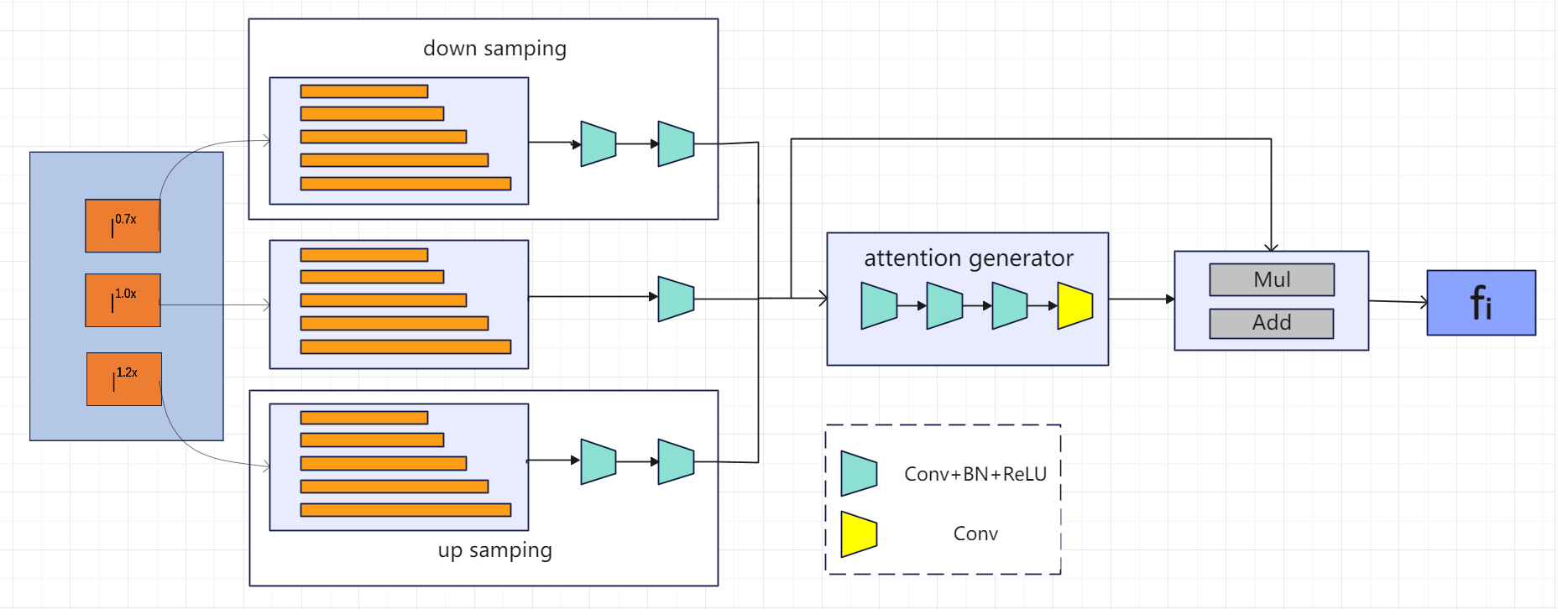}
    \caption{Illustration of the full scale merging module (FRM).}
    \label{fig:Figure 3}
\end{figure}
the three scales are intuitively set to 0.7x, 1.0x, 1.2x of the height and weight. The feature encoder is constituted by the commonly-used ResNeXt101~\cite{he2016deep} where the structure after “layer4” is removed. For more details, there will be five groups of features after the feature encoder, whose channels are  2048, 1024, 512, 256 and 64, respectively and whose heights and width are 1/32, 1/16, 1/8, 1/4, 1/2 of the original input size, respectively. Next, these groups of features are fed successively to the fine-rescaling and merging module (FRM) and the hierarchical channel-down decoder (HCDD) for subsequent processing.  

\subsection{Fine-Rescaling and Merging Module (FRM)}
While nearly every object in a given image maintains a consistent spatial relationship with other objects, which remains unchanged when the image is rescaled, the identification of edges, crucial for detecting concealed objects, can be significantly affected by rescaling, particularly in the case of low resolution and high levels of image noise in the test pictures. This impact becomes particularly evident when dealing with confusing glass-like objects and those with indistinct edges. To address this problem, we introduce the Fine-Rescaling and Merging module (\textbf{FRM}).

The attention-based FRM aims to weight and combine scale-specific information, as shown in Figure~\ref{fig:Figure 3}. Following~\cite{pang2022zoom}, the scale merging module is composed of several units and layers which implement filtering and aggregation to self-adaptivedly highlight the expression of different scales. The original scales (1.0x) is resized to 0.7 and 1,2 times of the lengths to generate relatively even input scales. After that, there are three groups of features belonging to 0.7x, 1.0x, and 1.2x, respectively. For \(f^{1.2}_i\), we further extract features by two stacked ``Conv-BN-ReLU" layers with different receptive fields and then use a mixed addition of ``max-pooling" and ``average-pooling" to down-sample it, which helps to conserve the meaningful and various responses for glass-like objects in high-resolution features. For \(f^{0.7}_i\), we further extract features by two stacked ``Conv-BN-ReLU" layers with the same receptive fields and up-sample it by bi-linear interpolation. For \(f^{1.0}_i\), it goes through a  ``Conv-BN-ReLU" layers. And these features are then fed into an attention generator stacked by three ``Conv-BN-ReLU" layers and one convolutional layer. 
\begin{figure}[ht]
    \centering
    \includegraphics[width=1.0\linewidth,height=3.5cm]{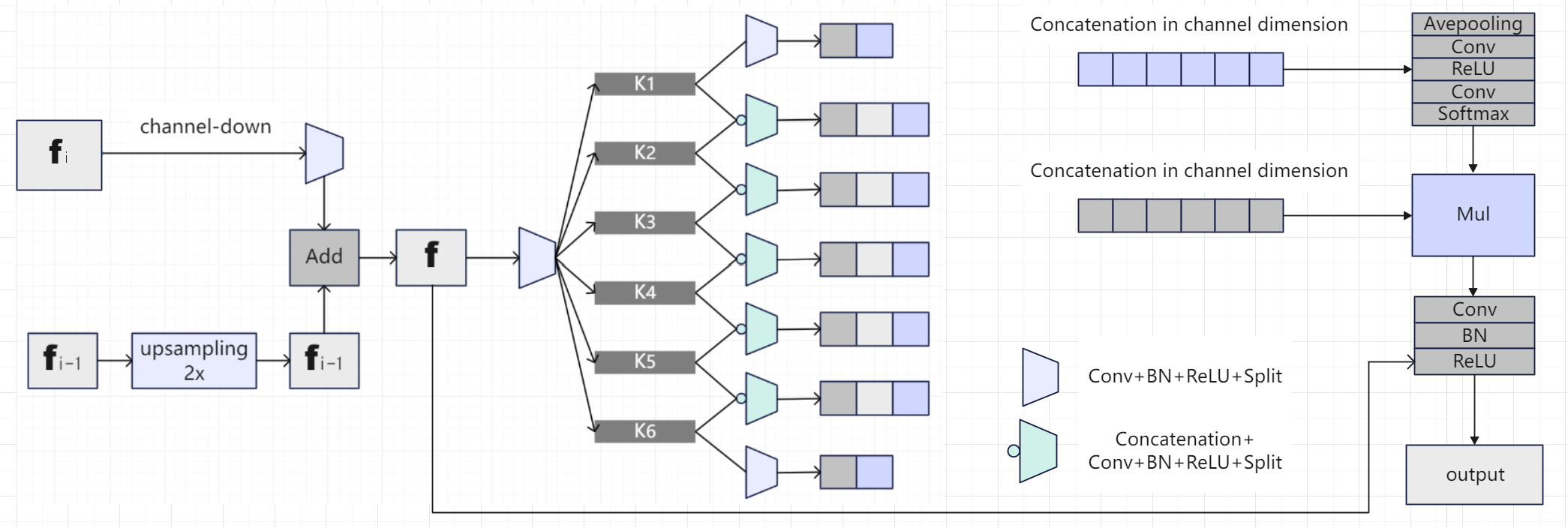}
    \caption{Illustration of the hierarchical channel-down unit (HCDU) .}
    \label{fig:Figure 4}
\end{figure}
A three-channel feature map is calculated through these layers and after a softmax activation layer, the attention map \(A^k\)(k \(\in\) \{0.7, 1.0, 1.2\}) corresponding to each scale can be obtained and used as respective weights for the final integration. The process is formulated as:
\begin{equation}
f_i^{0.7} = \zeta_1(f_i^{0.7}),  f_i^{1.0} = \zeta_2(f_i^{1.0}), 
 f_i^{1.2} = \zeta_3(f_i^{1.2}), 
\end{equation}
\begin{equation}
A_i = softmax(\xi([f_i^{1.2},f_i^{1.0},f_i^{0.7}],\theta))
\end{equation}
\begin{equation}
f_i = A_i^{1.2}*f_i^{1.2}+A_i^{1.0}*f_i^{1.0}+A_i^{0.7}*f_i^{0.7}
\end{equation} 
where \(\zeta_1\) indicates the stacked ``Conv-BN-ReLU" layers and the down-sample structure, \(\zeta_2\) is the stacked “Conv-BN-ReLU” layers, \(\zeta_3\) represents the stacked “Conv-BN-ReLU” layers and bi-linear interpolation, \(\xi\) denotes the stacked “Conv-BN-ReLU” layers in the attention generator, \(\theta\) is the parameters of these layers, and [] indicates input features are processed after the concatenation operation. 

\subsection{Hierarchical Channel-Down Decoder (HCDD)}
As multi-scales contain consistent compositions and accordant structures, after \textbf{FRM} integrates the multi-scale features into the original scale, the spatial and semantic relationships are highlighted in the feature representations. Similar to most cases, different channels also contain differentiated semantics. Thus, it is essential to mine valuable clues contained in different channels. Motivated by ~\cite{pang2022zoom}, we introduce Hierarchical Channel-Down Units \textbf{(HCDUs)} to interact and reorganise the critical information in different channels in a pyramid way, which strengthen features from coarse-grained group-wise iteration to fine-grained channel-wise modulation in the decoder, as shown in Fig.~\ref{fig:Figure 4}. The input 
 \(f_{i}\) of \(HCDU_{i}\) is the fused feature \(f_{6-i}\) from FRM and the input \(f_{i-1}\) is the output of \(HCDU_{i-1}\). We first reduce the channel of \(f_{i}\) by a stacked 3x3 “Conv-BN-ReLU" layer, up-sample  \(f_{i}\) by the direct bi-linear interpolation and add the two features as $f$:
 \begin{equation}
f = \delta(f_{i})+f_{i-1}^{2.0X}
\end{equation}
where $\delta$ represents the stacked “Conv-BN-ReLU" layer. 

\begin{figure}[ht]
    \centering
    \includegraphics[width=1\linewidth,height=4.0cm]{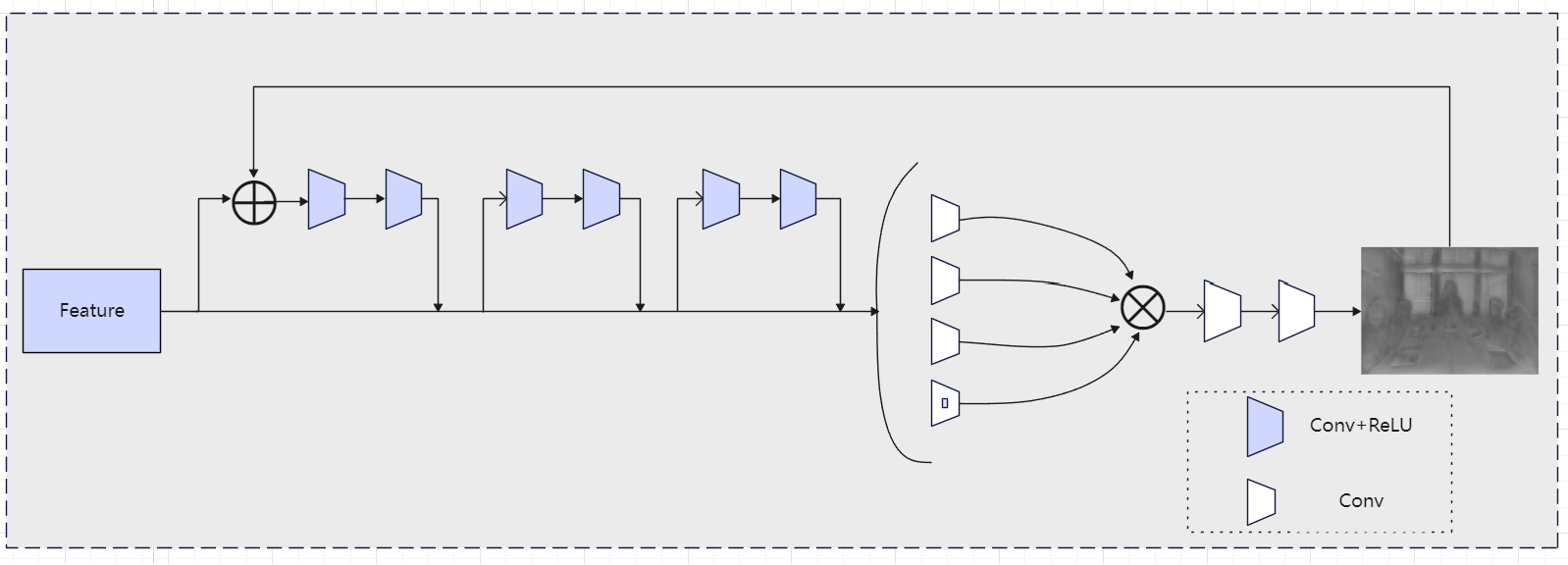}
    \caption{Illustration of the primary prediction guiding module (\textbf{PPG}).}
    \label{fig:Figure 5}
\end{figure}

To succinctly describe our unit, we define the stacked “Concatenation-Conv-BN-ReLU-Split” layers as CCBRS. The CCBRS processing the concatenation of the input two features is composed of a 3x3 convolutional layer, a batch normalization layer, a ReLU layer and a split layer breaking the feature into three or two chunks along the channel dimension. First, we expand the channels of $f$ six times by a 1x1 “Conv-BN-ReLU" layer and split $f$ into six chunks along the channel dimension $(k1,k2,k3,k4,k5,k6)$ shown in Fig.~\ref{fig:Figure 4}. Then the six features are successively processed in pair-wise groups through \textbf{CCBRS} and the unit processes the first and the last feature additionally by \textbf{CCBRS} without concatenation. The feature interaction is like a pyramid, as depicted in Fig.~\ref{fig:Figure 4}. Every group will generate two or three sub-features, the unit concatenates the last sub-feature in every sub-features group as the weighted feature, denoted by tensor \textbf{M}. The feature then goes through an average-pooling layer, a ReLU layer between two 1*1 convolutional layers and an activation layer to generate the weight feature map. Afterwards,the unit concatenates the first sub-feature in every sub-feature group, denoted by tensor \textbf{T} and multiple the tensor \textbf{T} by the weight feature map. The final output feature $\Tilde{f}$ can be obtained after a 3*3 convolutional layer, a batch normalization layer and a ReLU layer. The whole process can be formulated as:
 \begin{equation}
\Tilde{f} = \mathcal{R}eLU((f+\mathcal{N}(\mathcal{C}(\Psi(M) \cdot T))))
\end{equation}
where ReLU, $\mathcal{N}$, $\mathcal{C}$ represent the activation layer, the normalization layer and the convolutional layer, respectively and $\Psi$ denotes the weight generator layer.

The Hierarchical Channel-down Decoder (HCDD) contains five HCDUs. A single-channel logits map can be obtained after we integrate the five fused features with HCDUs and process the integrated features by a 1x1 convolutional layer. A coarse confidence map $P$ that somewhat highlights the glass-like objects can be then generated by a sigmoid function, as shown in Fig.~\ref{fig:Figure 2}.

\subsection{Primary Prediction Guiding Module (PPG)}

As it is obvious that the initial prediction is an important reference and contains  valuable clues about the rough position of the objects, we want our model to be able to guide further predictions with the original results. Thus, we introduce the primary prediction guiding module (PPG) to refine our prediction for detecting the severely confusing glass (mirror or glass baffle) and eliminating the uncertain regions in our segmentation map mainly due to vague boundaries, which is similar to the behaviour of human beings of checking more to confirm the structure and the edge.

Specifically, as shown in Fig.~\ref{fig:Figure 5}. The input feature of the iterative refinement module is the output feature of the HCDD. The final generated logits map can be represented as: 
 \begin{equation}
\mathcal{M}_t = \mathcal{F}(x, \mathcal{M}_{t-1})
\end{equation}
\(\mathcal{M}\) represents the output of the module, $t$ is the number of times and $x$ is the the ouput of the HCDD, which is unchanged during the refinement process. The function \(\mathcal{F}\) can be described as follow : in the first residual block, the feature $x$ concatenates with the logits map predicted in the last refinement process, and goes through the two 3x3 convolutional and ReLU layers (the first one has 33 filters and the second one has 32 filters) and is added with the original input feature. Output feature is denoted by \textbf{k1}. For the other two residual blocks, they are composed of two 3x3 convolutional and ReLU layers with 32 filters (the processed \textbf{k1} is denoted by \textbf{k2}) and \textbf{k2} is added with the original input feature \textbf{k1} (output feature is denoted by \textbf{k3}). After that, we apply the Atrous Spatial Pyramid Pooling module (ASPP) proposed in Deeplab V3~\cite{chen2017deeplab} to capture multi-scale information containing in \textbf{k3}. The module ASPP consists of four parallel branches that include three 3 × 3 convolutions with atrous rates of 6, 12, and 18, respectively and a 1 × 1 convolution. The 1 × 1 convolution is operated on the image-level feature which is achieved by global average pooling. The output features from 4 branches are concatenated and fused by another 1 × 1 convolution with 128 filters. The final single-channel logits map can be obtained after another 1x1 convolutional layer with 32 filters. This primary prediction guiding module implements a rather coarse-grained to fine-grained process, to preserve the critical semantic clues and further detect the targeted regions.

\begin{figure*}[ht]
    \centering
    \includegraphics[width=1.0\linewidth,height=5cm]{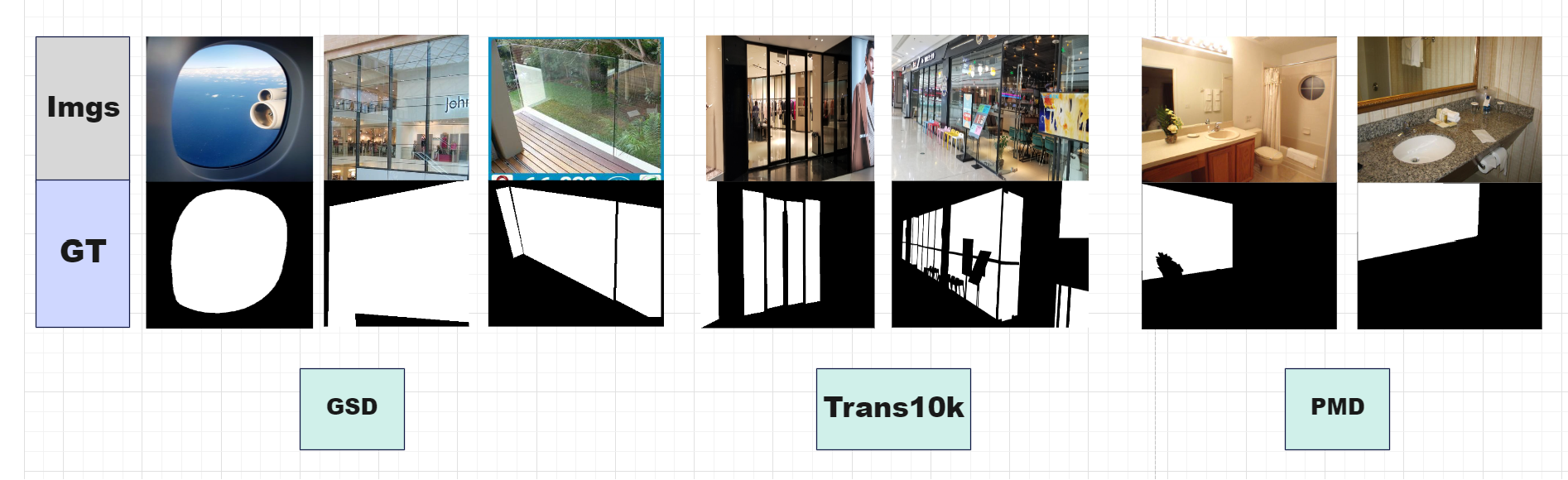}
    \caption{Examples of GSD, Trans10k and PMD dataset. Trans10k consists of 10,428 images with three categories: things, stuff and background. GSD contains 4,098 glass images, covering a diversity of indoor and outdoor scenes. PMD is a large-scale mirror dataset, containing a variety of real-world images that cover diverse scenes and common objects, making it much closer to practical applications.}
    \label{fig:Figure 6}
\end{figure*}

\subsection{Loss Functions}
The binary cross entropy loss (BCE) which is widely used in image segmentation tasks is adopted as a part of our loss function and its mathematical form is $$l_{BCE}^{i,j} = -g_{i,j}logp_{i,j}-(1-g_{i,j})log(1-p_{i,j})$$ where $g_{i,j}$ \(\in\) \{0, 1\} and $p_{i,j}$ \(\in\) $[0,1]$ denote the ground truth and the predicted value at position $(i,j)$, respectively.

To eliminate the uncertainties of the segmentation map in the iterative refinement module and generate high-confidence results, we use an auxiliary loss function of the BCE, i.e., the uncertainty-aware loss (UAL) proposed in \cite{pang2022zoom}. In the final probability map of the camouflaged object, the pixel value range is $
[0,1]$, where $0$ means the pixel belongs to the background, and $1$ means it belongs to the camouflaged object. Therefore, the closer the predicted value is to $0.5$, the more uncertain the determination about the property of the pixel is. The UAL maximizes at $x = 0.5$ and minimizes at $x = 0$ or $x = 1$ to optimize the ambiguity of the prediction. the UAL is formulated as $l_{UAL}^{i,j} = 1 - |2p_{i,j}-1|^2$.

So the final loss function is as follows:
 \begin{equation}
L = L_{BCE} + 1.5*\lambda * L_{UAL}
\end{equation}
where \(\lambda\) ranged from zero to one is the balance coefficient which utilizes an increasing cosine strategy.

\section{Experimental Results}
\subsection{Experimental Settings}
\textbf{Datasets.}  We evaluate the proposed method on three widely used glass and mirror datasets i.e., Trans10k~\cite{xie2020segmenting}, GSD~\cite{lin2021rich} and PMD~\cite{lin2020progressive}. Trans10k is a large-scale transparent object segmentation dataset, consisting of 10,428 images with three categories: things, stuff and background. Images are divided into 5,000, 1,000 and 4,428 images for training, validation and test, respectively. GSD is a medium-scale glass segmentation dataset containing 4,098 glass images, covering a diversity of indoor and outdoor scenes. All the data are randomly split
into a training set with 3,285 images and a test set with 813
images. PMD is another large-scale mirror dataset containing 5,096 training images and 571 test images. It has a variety of real-world images that cover
diverse scenes and common objects, making it much closer to practical applications.
Unlike existing methods, we trained our model with respect to each dataset under consistent settings, while the existing glass segmentation models must be trained on different settings with respect to varied datasets.

\textbf{Evaluation Criteria.}
We apply the following evaluation metrics to assess the performance of our model: mean Intersection over Union (mIoU), Mean Absolute Error (mAE), and mean Balance Error Rate (mBER). The mIoU is commonly employed to determine the ratio of true positive predictions. The mBER offers a more inclusive assessment of error rates, accounting for sample imbalance and mAE measures the absolute prediction error of the segmentation map.

 \textbf{Implementation Details.} The proposed MGNet is implemented with PyTorch. The encoder is initialized with the parameters of pre-trained ResNeXt101 and the remaining parts are randomly initialized. And SGD with momentum 0.9 and weight decay 0.0005 is chosen as the optimizer. The learning rate is initialized to 0.08 and follows a linear warmup and linear decay strategy. The entire model is trained for 32 epochs with a batch size of 12 in an end-to-end manner on an NVIDIA Tesla V100-SXM2-32GB GPU. During training and inference, the input scale is 384 × 384. Random flipping and rotating are employed to augment the training data.

\subsection{Comparisons with State-of-the-art}
To demonstrate the effectiveness of the proposed method, we compare it with several state-of-the-art glass-like object segmentation models, including mirror detection and transparent object detection. All the predictions of baselines are either provided by the authors or generated by open source models retrained .

\textbf{Quantitative Comparison.} Tab.~\ref{Table 1} summarizes the quantitative results of our proposed method against 10 baselines on the challenging glass surface benchmark dataset under three evaluation metrics.  Tab.~\ref{Table 2} presents the quantitative results of our proposed method against 8 counterparts on the demanding mirror benchmark dataset under two evaluation metrics. We can see that our model has outperformed all the state-of-the-art baselines both in IoU and MAE and has comparable performance in BER. 

\textbf{Qualitative Evaluation.} Our method also achieves excellent qualitative results which are exhibited in the Figure ~\ref{fig:Figure 1} and ~\ref{fig:Figure 7}.

\begin{table}[]
\caption{Quantitative comparison with 11 SOTA methods on GSD benchmark dataset. Notes ↑ / ↓ denote the larger/smaller is better, respectively. “–” is not available. The best and second best are highlighted in \textcolor{red}{red} and \textcolor{blue}{blue} respectively.}
\label{Table 1}
\resizebox{\linewidth}{33mm}{
\begin{tabular}{l|l|lll|}
\hline
\multirow{2}{*}{\textbf{Methods}} & \multirow{2}{*}{Publication Venues} & \multicolumn{3}{l|}{\textbf{GSD (glass surface dataset)}}        \\ \cline{3-5} 
                                  &                               & \multicolumn{1}{l|}{IoU{$\uparrow$}}   & \multicolumn{1}{l|}{MAE{$\downarrow$}}   & BER{$\downarrow$}   \\ \hline
PSPNet~\cite{zhao2017pyramid}                            & CVPR2017                      & \multicolumn{1}{l|}{70.26} & \multicolumn{1}{l|}{0.110} & 10.66 \\ \hline
BDRAR~\cite{zhu2018bidirectional}                             & ECCV2018                      & \multicolumn{1}{l|}{75.91} & \multicolumn{1}{l|}{0.081} & 8.61  \\ \hline
BASNet~\cite{qin2019basnet}                            & CVPR2019                      & \multicolumn{1}{l|}{69.79} & \multicolumn{1}{l|}{0.106} & 13.54 \\ \hline
MINet~\cite{pang2020multi}                             & CVPR2020                      & \multicolumn{1}{l|}{77.29} & \multicolumn{1}{l|}{0.077} & 9.54  \\ \hline
SINet\cite{fan2020camouflaged}                             & CVPR2020                      & \multicolumn{1}{l|}{77.04} & \multicolumn{1}{l|}{0.077} & 9.25  \\ \hline
GDNet~\cite{mei2020don}                             & CVPR2020                      & \multicolumn{1}{l|}{79.01} & \multicolumn{1}{l|}{0.069} & 7.72  \\ \hline
TransLab\cite{xie2020segmenting}                          & ECCV2020                      & \multicolumn{1}{l|}{74.05} & \multicolumn{1}{l|}{0.088} & 11.35 \\ \hline
GSDNet~\cite{lin2021rich}                               & CVPR2021                      & \multicolumn{1}{l|}{83.64} & \multicolumn{1}{l|}{0.055} &  \textcolor{red}{6.12}  \\ \hline
PGSNet~\cite{mei2022glass}                            & CVPR2022                      & \multicolumn{1}{l|}{\textcolor{blue}{83.65}} & \multicolumn{1}{l|}{\textcolor{blue}{0.054}} & \textcolor{blue} {6.25}  \\ \hline
SAM~\cite{kirillov2023segment}                               & ICCV2023                      & \multicolumn{1}{l|}{50.60} & \multicolumn{1}{l|}{0.213} & 23.91 \\ \hline
\textbf{MGNet}(Ours)                             & \multicolumn{1}{c|}{---}                             & \multicolumn{1}{l|} {\textcolor{red}{84.08}} & \multicolumn{1}{l|} {\textcolor{red}{0.051}} & 6.32  \\ \hline
\end{tabular}
}
\end{table}

\subsection{Ablation Studies}
To validate the effectiveness of the proposed modules in our model, we have also performed the following ablation studies on these benchmarks.

\textbf{Effectiveness of FRM}. Integrating multiple scales of input pictures has been found useful for understanding the complex spatial relationship of the glass-like objects with other objects. The FRM also enables our model to adjust attention to different scales with respect to each pixel. To confirm its effectiveness, we delete the module from MGNet, preserving other modules without any modification, train it under the same settings and test it on two datasets (GSD 
 and PMD), with the results shown in Table ~\ref{Table 4}. It can be easily seen that the model has made great progress on all the metrics after incorporating the FRM to integrate the critical semantics of different input scales.

\begin{figure*}[ht]
    \centering
    \includegraphics[width=1.0\linewidth,height=10cm]{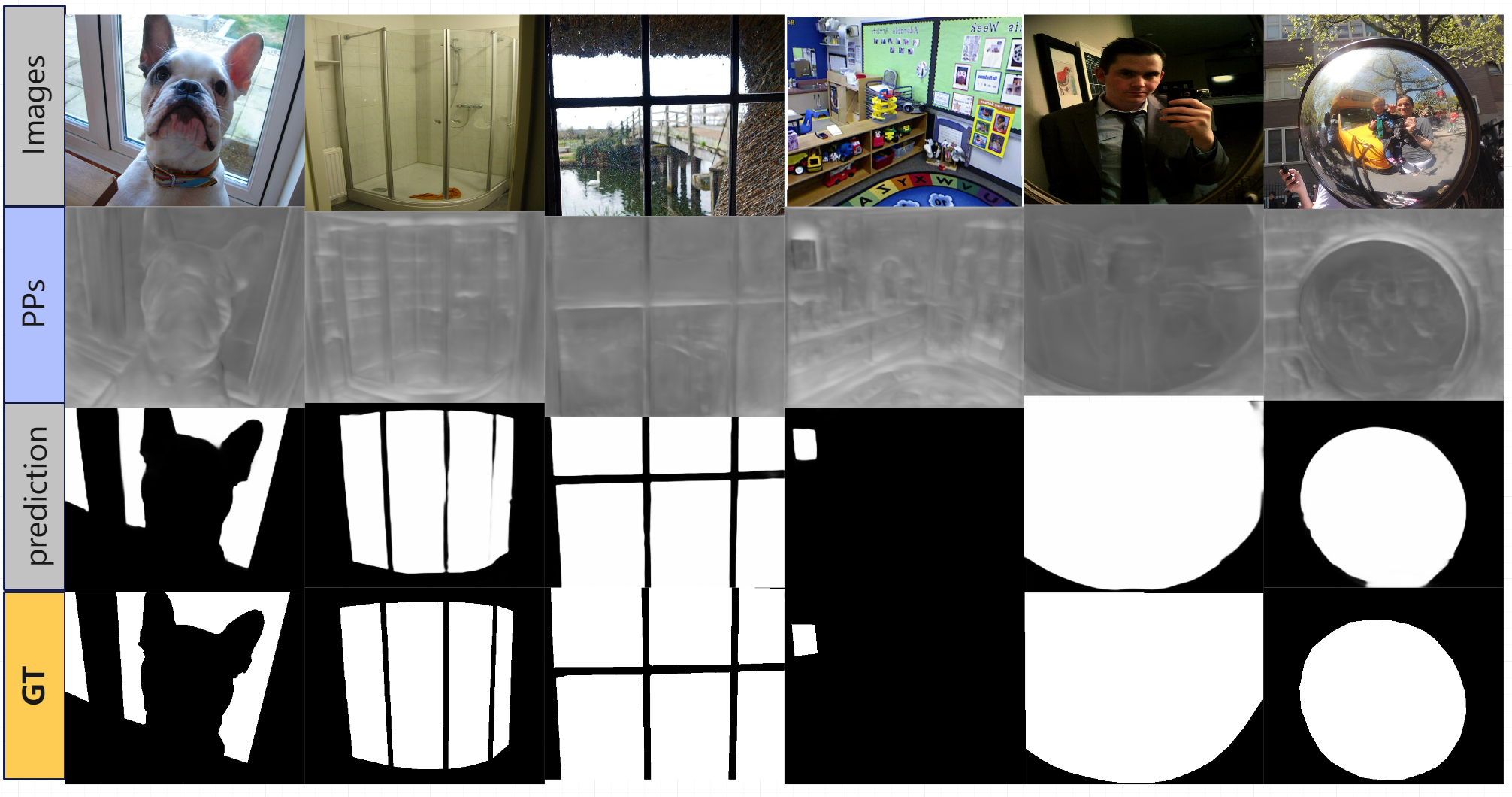}
    \caption{\textbf{Visualization} of the segmentation process of our model. The \textbf{PPs} row is the visualization of the primary prediction without refinement. The \textbf{predition} row is the final results generated by PPG. }
    \label{fig:Figure 7}
\end{figure*}

 \textbf{Effectiveness of PPG}. Unlike existing iterative refinement strategies, our refinement strategy aims to implement a coarse-to-fine segmentation process and capture the structure and edge of the glass-like objects with the primary prediction, which is also the process of eliminating the uncertainty in the segmentation map. The segmentation process can be visualized in Fig.~\ref{fig:Figure 7}. It is obvious that our model first generates a coarse logits map which slightly highlights the glass-like objects but still preserves the most critical semantic clues and leaves many uncertain regions mainly due to the vague boundaries. The results demonstrate convincingly that our model successfully implement the idea of refinement with the primary prediction to get more accurate and complete answers. Then, based on this coarse logits map, the segmentation map are refined and becomes more and more regionally distinct. To confirm its effectiveness of improving the performance of our model, we uninstall the primary prediction guiding module (PPG) from MGNet. It can be seen in Table ~\ref{Table 4} that with PPG, the model has obvious performance boost of the weighted mIoU, MAE and BER on the two datasets (GSD and PMD).

\begin{table}[]
\caption{ Quantitative comparison with 8 SOTA methods on PMD benchmark dataset. Notes ↑ / ↓ denote the larger/smaller is better, respectively. “–” is not available. The best and second best are highlighted in \textcolor{red}{red} and \textcolor{blue}{blue} respectively.}
\label{Table 2}
\resizebox{\linewidth}{30mm}{
\begin{tabular}{l|l|ll|}
\hline
\multirow{2}{*}{Methods} & \multirow{2}{*}{Publication Venues} & \multicolumn{2}{l|}{PMD (mirror dataset)} \\ \cline{3-4} 
                         &                               & {IoU{$\uparrow$}}                 & {MAE{$\downarrow$}}                \\ \hline
EGNet~\cite{zhao2019egnet}                    & ICCV2019                      & 60.17               & \textcolor{blue}{0.036}              \\
BDRAR~\cite{zhu2018bidirectional}                    & ECCV2018                      & 58.43               & 0.043              \\
PSPNet~\cite{zhao2017pyramid}                   & CVPR2017                      & 60.44               & 0.039              \\
DeepLabv3+~\cite{chen2018encoder}               & ECCV2018                      & 64.08               & 0.040              \\
PMDNet~\cite{lin2020progressive}                   & CVPR2020                      & 62.40               & 0.087              \\
MirrorNet~\cite{yang2019my}                & ICCV2019                      & 62.50               & 0.041              \\
Guan et al.~\cite{guan2022learning}              & CVPR2022                      & \textcolor{blue}{66.84}               & 0.049              \\
SAM~\cite{kirillov2023segment}                      & ICCV2023                      & 64.75               & 0.052              \\ \hline
MGNet                    & \multicolumn{1}{c|}{---}                                & \textcolor{red}{68.77}               & \textcolor{red}{0.028}              \\ \hline
\end{tabular}
}
\end{table}

\textbf{Uncertainty-aware Loss}. In the final probability map of the glass-like objects, the pixel value range is $[0,1]$, where $0$ means the pixel belongs to the background, and $1$ represents it belongs to the camouflaged object. The closer the predicted value is to $0.5$, the more uncertain the determination about the property of the pixel is. Also with the purpose of enabling our model to eliminate the uncertainty during refining, the UAL should maximize at $x = 0.5$ and minimize at $x = 0$ or $x = 1$ to optimize the ambiguity of the prediction. To confirm its effectiveness, we delete the UAL 
\begin{table}[h]
\caption{Quantitative comparison between the complete model (MGNet) and ablative models on two benchmark datasets (GSD~\cite{lin2021rich} and PMD~\cite{lin2020progressive}). And \textbf{B}, \textbf{F}, \textbf{H}, \textbf{P} and \textbf{U} denote backbone, FRM, HCDD, PPG and UAL, respectively. “+” means addition of the module. The results of our complete model (MGNet) are \textbf{bolded}.}
\label{Table 4}
\resizebox{\linewidth}{15mm}{
\begin{tabular}{l|lll|ll|}
\hline
\multirow{2}{*}{Settings} & \multicolumn{3}{l|}{GSD}                        & \multicolumn{2}{l|}{PMD}        \\ \cline{2-6} 
                          & {IoU{$\uparrow$}}           &{MAE{$\downarrow$}}          & BER{$\downarrow$}           & {IoU{$\uparrow$}}             &{MAE{$\downarrow$}}             \\ \hline
B+H+P                     & 75.74          & 0.082          & 9.04          & 60.95          & 0.042          \\
B+H+F                     & 77.13          & 0.077          & 8.46          & 61.89          & 0.040          \\
B+H+F+P                 & 80.12          & 0.059          & 7.23          & 64.22          & 0.037          \\ \hline
\textbf{B+H+F+P+U (our setting)} & \textbf{84.08} & \textbf{0.051} & \textbf{6.32} & \textbf{68.77} & \textbf{0.028} \\ \hline
\end{tabular}
}
\end{table}
\noindent from our model, and train with the same settings. The results in Table ~\ref{Table 4} indicate that the performance of our model is generally enhanced after adding the UAL to the loss function, justifying its efficacy. 

Furthermore, for another transparent object instances dataset Trans10k, Table~\ref{Table 3} demonstrates the quantitative results of our proposed MGNet with all modules installed, showing that our model can achieve good performance on this dataset. Different from model comparison, the evaluation on this dataset is to test the model's capability to detect the presence of glass-like objects. Specifically, all ground truth annotations are binary, with areas containing glass-like objects marked in white and the background marked in black.   
\begin{table}[]
\caption{Quantitative test of our model MGNet on Trans10k benchmark dataset. Results show that our model has good performance on segmenting transparent objects.}
\label{Table 3}
\resizebox{\linewidth}{9mm}{\begin{tabular}{l|l|lll|}
\hline
\multirow{2}{*}{Methods} & \multirow{2}{*}{Settings} & \multicolumn{3}{c|}{Trans10K}      \\ \cline{3-5} 
                         &                               & \multicolumn{1}{l|}{IoU{$\uparrow$}}    & \multicolumn{1}{l|}{MAE{$\downarrow$}}   & BER{$\downarrow$}  \\ \hline
MGNet                    & \multicolumn{1}{c|}{\textbf{B+H+F+P+U}}                             & \multicolumn{1}{l|}{89.97} & \multicolumn{1}{l|}{0.030} & 4.05 \\ \hline
\end{tabular}
}
\end{table}

\section{Conclusion}
In this paper, we address the challenging problem of glass-like object segmentation with the proposed MGNet, which contains three novel modules of the fine-rescaling and merging (\textbf{FRM}) for integrating multi-scales features, the hierarchical channel-down decoder (\textbf{HCDD}) for further mining valuable clues contained in different channels, and the primary prediction guiding module (\textbf{PPG}) for refining more overall structures and detailed information. Extensive experiments show that our model can achieve superior performance on Trans10k, GSD and PMD datasets with respect to state-of-the-art counterparts. And the future work will explore more advanced techniques for extracting more nuanced and richer features for better segmentation.

{\small
\nocite{*}
\bibliographystyle{ieee_fullname}
\bibliography{egbib}

\begin{thebibliography}{10}\itemsep=-1pt

\bibitem{adelson1984pyramid}
Edward~H Adelson, Charles~H Anderson, James~R Bergen, Peter~J Burt, and Joan~M Ogden.
\newblock Pyramid methods in image processing.
\newblock {\em RCA engineer}, 29(6):33--41, 1984.

\bibitem{chen2017deeplab}
Liang-Chieh Chen, George Papandreou, Iasonas Kokkinos, Kevin Murphy, and Alan~L Yuille.
\newblock Deeplab: Semantic image segmentation with deep convolutional nets, atrous convolution, and fully connected crfs.
\newblock {\em IEEE transactions on pattern analysis and machine intelligence}, 40(4):834--848, 2017.

\bibitem{chen2018encoder}
Liang-Chieh Chen, Yukun Zhu, George Papandreou, Florian Schroff, and Hartwig Adam.
\newblock Encoder-decoder with atrous separable convolution for semantic image segmentation.
\newblock In {\em Proceedings of the European conference on computer vision (ECCV)}, pages 801--818, 2018.

\bibitem{chen2018reverse}
Shuhan Chen, Xiuli Tan, Ben Wang, and Xuelong Hu.
\newblock Reverse attention for salient object detection.
\newblock In {\em Proceedings of the European conference on computer vision (ECCV)}, pages 234--250, 2018.

\bibitem{fan2020camouflaged}
Deng-Ping Fan, Ge-Peng Ji, Guolei Sun, Ming-Ming Cheng, Jianbing Shen, and Ling Shao.
\newblock Camouflaged object detection.
\newblock In {\em Proceedings of the IEEE/CVF conference on computer vision and pattern recognition}, pages 2777--2787, 2020.

\bibitem{fan2023rfenet}
Ke Fan, Changan Wang, Yabiao Wang, Chengjie Wang, Ran Yi, and Lizhuang Ma.
\newblock Rfenet: Towards reciprocal feature evolution for glass segmentation.
\newblock {\em arXiv preprint arXiv:2307.06099}, 2023.

\bibitem{gao2020highly}
Shang-Hua Gao, Yong-Qiang Tan, Ming-Ming Cheng, Chengze Lu, Yunpeng Chen, and Shuicheng Yan.
\newblock Highly efficient salient object detection with 100k parameters.
\newblock In {\em European Conference on Computer Vision}, pages 702--721. Springer, 2020.

\bibitem{guan2022learning}
Huankang Guan, Jiaying Lin, and Rynson~WH Lau.
\newblock Learning semantic associations for mirror detection.
\newblock In {\em Proceedings of the IEEE/CVF Conference on Computer Vision and Pattern Recognition}, pages 5941--5950, 2022.

\bibitem{han2023segment}
Dongsheng Han, Chaoning Zhang, Yu Qiao, Maryam Qamar, Yuna Jung, SeungKyu Lee, Sung-Ho Bae, and Choong~Seon Hong.
\newblock Segment anything model (sam) meets glass: Mirror and transparent objects cannot be easily detected.
\newblock {\em arXiv preprint arXiv:2305.00278}, 2023.

\bibitem{he2015spatial}
Kaiming He, Xiangyu Zhang, Shaoqing Ren, and Jian Sun.
\newblock Spatial pyramid pooling in deep convolutional networks for visual recognition.
\newblock {\em IEEE transactions on pattern analysis and machine intelligence}, 37(9):1904--1916, 2015.

\bibitem{he2016deep}
Kaiming He, Xiangyu Zhang, Shaoqing Ren, and Jian Sun.
\newblock Deep residual learning for image recognition.
\newblock In {\em Proceedings of the IEEE conference on computer vision and pattern recognition}, pages 770--778, 2016.

\bibitem{ji2022dmra}
Wei Ji, Ge Yan, Jingjing Li, Yongri Piao, Shunyu Yao, Miao Zhang, Li Cheng, and Huchuan Lu.
\newblock Dmra: Depth-induced multi-scale recurrent attention network for rgb-d saliency detection.
\newblock {\em IEEE Transactions on Image Processing}, 31:2321--2336, 2022.

\bibitem{jia2022segment}
Qi Jia, Shuilian Yao, Yu Liu, Xin Fan, Risheng Liu, and Zhongxuan Luo.
\newblock Segment, magnify and reiterate: Detecting camouflaged objects the hard way.
\newblock In {\em Proceedings of the IEEE/CVF Conference on Computer Vision and Pattern Recognition}, pages 4713--4722, 2022.

\bibitem{kirillov2023segment}
Alexander Kirillov, Eric Mintun, Nikhila Ravi, Hanzi Mao, Chloe Rolland, Laura Gustafson, Tete Xiao, Spencer Whitehead, Alexander~C Berg, Wan-Yen Lo, et~al.
\newblock Segment anything.
\newblock {\em arXiv preprint arXiv:2304.02643}, 2023.

\bibitem{li2015visual}
Guanbin Li and Yizhou Yu.
\newblock Visual saliency based on multiscale deep features.
\newblock In {\em Proceedings of the IEEE conference on computer vision and pattern recognition}, pages 5455--5463, 2015.

\bibitem{lin2021rich}
Jiaying Lin, Zebang He, and Rynson~WH Lau.
\newblock Rich context aggregation with reflection prior for glass surface detection.
\newblock In {\em Proceedings of the IEEE/CVF Conference on Computer Vision and Pattern Recognition}, pages 13415--13424, 2021.

\bibitem{lin2020progressive}
Jiaying Lin, Guodong Wang, and Rynson~WH Lau.
\newblock Progressive mirror detection.
\newblock In {\em Proceedings of the IEEE/CVF Conference on Computer Vision and Pattern Recognition}, pages 3697--3705, 2020.

\bibitem{lin2017feature}
Tsung-Yi Lin, Piotr Doll{\'a}r, Ross Girshick, Kaiming He, Bharath Hariharan, and Serge Belongie.
\newblock Feature pyramid networks for object detection.
\newblock In {\em Proceedings of the IEEE conference on computer vision and pattern recognition}, pages 2117--2125, 2017.

\bibitem{lindeberg1998feature}
Tony Lindeberg.
\newblock Feature detection with automatic scale selection.
\newblock {\em International journal of computer vision}, 30:79--116, 1998.

\bibitem{lindeberg2013scale}
Tony Lindeberg.
\newblock {\em Scale-space theory in computer vision}, volume 256.
\newblock Springer Science \& Business Media, 2013.

\bibitem{liu2018picanet}
Nian Liu, Junwei Han, and Ming-Hsuan Yang.
\newblock Picanet: Learning pixel-wise contextual attention for saliency detection.
\newblock In {\em Proceedings of the IEEE conference on computer vision and pattern recognition}, pages 3089--3098, 2018.

\bibitem{liu2021samnet}
Yun Liu, Xin-Yu Zhang, Jia-Wang Bian, Le Zhang, and Ming-Ming Cheng.
\newblock Samnet: Stereoscopically attentive multi-scale network for lightweight salient object detection.
\newblock {\em IEEE Transactions on Image Processing}, 30:3804--3814, 2021.

\bibitem{long2015fully}
Jonathan Long, Evan Shelhamer, and Trevor Darrell.
\newblock Fully convolutional networks for semantic segmentation.
\newblock In {\em Proceedings of the IEEE conference on computer vision and pattern recognition}, pages 3431--3440, 2015.

\bibitem{luo2017non}
Zhiming Luo, Akshaya Mishra, Andrew Achkar, Justin Eichel, Shaozi Li, and Pierre-Marc Jodoin.
\newblock Non-local deep features for salient object detection.
\newblock In {\em Proceedings of the IEEE Conference on computer vision and pattern recognition}, pages 6609--6617, 2017.

\bibitem{mei2022glass}
Haiyang Mei, Bo Dong, Wen Dong, Jiaxi Yang, Seung-Hwan Baek, Felix Heide, Pieter Peers, Xiaopeng Wei, and Xin Yang.
\newblock Glass segmentation using intensity and spectral polarization cues.
\newblock In {\em Proceedings of the IEEE/CVF Conference on Computer Vision and Pattern Recognition}, pages 12622--12631, 2022.

\bibitem{mei2020don}
Haiyang Mei, Xin Yang, Yang Wang, Yuanyuan Liu, Shengfeng He, Qiang Zhang, Xiaopeng Wei, and Rynson~WH Lau.
\newblock Don't hit me! glass detection in real-world scenes.
\newblock In {\em Proceedings of the IEEE/CVF Conference on Computer Vision and Pattern Recognition}, pages 3687--3696, 2020.

\bibitem{pang2020hierarchical}
Youwei Pang, Lihe Zhang, Xiaoqi Zhao, and Huchuan Lu.
\newblock Hierarchical dynamic filtering network for rgb-d salient object detection.
\newblock In {\em Computer Vision--ECCV 2020: 16th European Conference, Glasgow, UK, August 23--28, 2020, Proceedings, Part XXV 16}, pages 235--252. Springer, 2020.

\bibitem{pang2022zoom}
Youwei Pang, Xiaoqi Zhao, Tian-Zhu Xiang, Lihe Zhang, and Huchuan Lu.
\newblock Zoom in and out: A mixed-scale triplet network for camouflaged object detection.
\newblock In {\em Proceedings of the IEEE/CVF Conference on computer vision and pattern recognition}, pages 2160--2170, 2022.

\bibitem{pang2020multi}
Youwei Pang, Xiaoqi Zhao, Lihe Zhang, and Huchuan Lu.
\newblock Multi-scale interactive network for salient object detection.
\newblock In {\em Proceedings of the IEEE/CVF conference on computer vision and pattern recognition}, pages 9413--9422, 2020.

\bibitem{qin2019basnet}
Xuebin Qin, Zichen Zhang, Chenyang Huang, Chao Gao, Masood Dehghan, and Martin Jagersand.
\newblock Basnet: Boundary-aware salient object detection.
\newblock In {\em Proceedings of the IEEE/CVF conference on computer vision and pattern recognition}, pages 7479--7489, 2019.

\bibitem{ren2015faster}
Shaoqing Ren, Kaiming He, Ross Girshick, and Jian Sun.
\newblock Faster r-cnn: Towards real-time object detection with region proposal networks.
\newblock {\em Advances in neural information processing systems}, 28, 2015.

\bibitem{ronneberger2015u}
Olaf Ronneberger, Philipp Fischer, and Thomas Brox.
\newblock U-net: Convolutional networks for biomedical image segmentation.
\newblock In {\em Medical Image Computing and Computer-Assisted Intervention--MICCAI 2015: 18th International Conference, Munich, Germany, October 5-9, 2015, Proceedings, Part III 18}, pages 234--241. Springer, 2015.

\bibitem{simonyan2014very}
Karen Simonyan and Andrew Zisserman.
\newblock Very deep convolutional networks for large-scale image recognition.
\newblock {\em arXiv preprint arXiv:1409.1556}, 2014.

\bibitem{wang2019salient}
Wenguan Wang, Shuyang Zhao, Jianbing Shen, Steven~CH Hoi, and Ali Borji.
\newblock Salient object detection with pyramid attention and salient edges.
\newblock In {\em Proceedings of the IEEE/CVF conference on computer vision and pattern recognition}, pages 1448--1457, 2019.

\bibitem{witkin1984scale}
Andrew Witkin.
\newblock Scale-space filtering: A new approach to multi-scale description.
\newblock In {\em ICASSP'84. IEEE International Conference on Acoustics, Speech, and Signal Processing}, volume~9, pages 150--153. IEEE, 1984.

\bibitem{wu2019cascaded}
Zhe Wu, Li Su, and Qingming Huang.
\newblock Cascaded partial decoder for fast and accurate salient object detection.
\newblock In {\em Proceedings of the IEEE/CVF conference on computer vision and pattern recognition}, pages 3907--3916, 2019.

\bibitem{xie2020segmenting}
Enze Xie, Wenjia Wang, Wenhai Wang, Mingyu Ding, Chunhua Shen, and Ping Luo.
\newblock Segmenting transparent objects in the wild.
\newblock In {\em Computer Vision--ECCV 2020: 16th European Conference, Glasgow, UK, August 23--28, 2020, Proceedings, Part XIII 16}, pages 696--711. Springer, 2020.

\bibitem{yan2021mirrornet}
Jinnan Yan, Trung-Nghia Le, Khanh-Duy Nguyen, Minh-Triet Tran, Thanh-Toan Do, and Tam~V Nguyen.
\newblock Mirrornet: Bio-inspired camouflaged object segmentation.
\newblock {\em IEEE Access}, 9:43290--43300, 2021.

\bibitem{yan2013hierarchical}
Qiong Yan, Li Xu, Jianping Shi, and Jiaya Jia.
\newblock Hierarchical saliency detection.
\newblock In {\em Proceedings of the IEEE conference on computer vision and pattern recognition}, pages 1155--1162, 2013.

\bibitem{yang2013saliency}
Chuan Yang, Lihe Zhang, Huchuan Lu, Xiang Ruan, and Ming-Hsuan Yang.
\newblock Saliency detection via graph-based manifold ranking.
\newblock In {\em Proceedings of the IEEE conference on computer vision and pattern recognition}, pages 3166--3173, 2013.

\bibitem{yang2021uncertainty}
Fan Yang, Qiang Zhai, Xin Li, Rui Huang, Ao Luo, Hong Cheng, and Deng-Ping Fan.
\newblock Uncertainty-guided transformer reasoning for camouflaged object detection.
\newblock In {\em Proceedings of the IEEE/CVF International Conference on Computer Vision}, pages 4146--4155, 2021.

\bibitem{yang2019my}
Xin Yang, Haiyang Mei, Ke Xu, Xiaopeng Wei, Baocai Yin, and Rynson~WH Lau.
\newblock Where is my mirror?
\newblock In {\em Proceedings of the IEEE/CVF International Conference on Computer Vision}, pages 8809--8818, 2019.

\bibitem{zeng2019towards}
Yi Zeng, Pingping Zhang, Jianming Zhang, Zhe Lin, and Huchuan Lu.
\newblock Towards high-resolution salient object detection.
\newblock In {\em Proceedings of the IEEE/CVF international conference on computer vision}, pages 7234--7243, 2019.

\bibitem{zhang2019canet}
Chi Zhang, Guosheng Lin, Fayao Liu, Rui Yao, and Chunhua Shen.
\newblock Canet: Class-agnostic segmentation networks with iterative refinement and attentive few-shot learning.
\newblock In {\em Proceedings of the IEEE/CVF conference on computer vision and pattern recognition}, pages 5217--5226, 2019.

\bibitem{zhang2021auto}
Miao Zhang, Tingwei Liu, Yongri Piao, Shunyu Yao, and Huchuan Lu.
\newblock Auto-msfnet: Search multi-scale fusion network for salient object detection.
\newblock In {\em Proceedings of the 29th ACM international conference on multimedia}, pages 667--676, 2021.

\bibitem{zhao2017pyramid}
Hengshuang Zhao, Jianping Shi, Xiaojuan Qi, Xiaogang Wang, and Jiaya Jia.
\newblock Pyramid scene parsing network.
\newblock In {\em Proceedings of the IEEE conference on computer vision and pattern recognition}, pages 2881--2890, 2017.

\bibitem{zhao2019egnet}
Jia-Xing Zhao, Jiang-Jiang Liu, Deng-Ping Fan, Yang Cao, Jufeng Yang, and Ming-Ming Cheng.
\newblock Egnet: Edge guidance network for salient object detection.
\newblock In {\em Proceedings of the IEEE/CVF international conference on computer vision}, pages 8779--8788, 2019.

\bibitem{zhao2019pyramid}
Ting Zhao and Xiangqian Wu.
\newblock Pyramid feature attention network for saliency detection.
\newblock In {\em Proceedings of the IEEE/CVF conference on computer vision and pattern recognition}, pages 3085--3094, 2019.

\bibitem{zhao2021automatic}
Xiaoqi Zhao, Lihe Zhang, and Huchuan Lu.
\newblock Automatic polyp segmentation via multi-scale subtraction network.
\newblock In {\em Medical Image Computing and Computer Assisted Intervention--MICCAI 2021: 24th International Conference, Strasbourg, France, September 27--October 1, 2021, Proceedings, Part I 24}, pages 120--130. Springer, 2021.

\bibitem{zheng2018detection}
Yunfei Zheng, Xiongwei Zhang, Feng Wang, Tieyong Cao, Meng Sun, and Xiaobing Wang.
\newblock Detection of people with camouflage pattern via dense deconvolution network.
\newblock {\em IEEE Signal Processing Letters}, 26(1):29--33, 2018.

\bibitem{zhou2020interactive}
Huajun Zhou, Xiaohua Xie, Jian-Huang Lai, Zixuan Chen, and Lingxiao Yang.
\newblock Interactive two-stream decoder for accurate and fast saliency detection.
\newblock In {\em Proceedings of the IEEE/CVF conference on computer vision and pattern recognition}, pages 9141--9150, 2020.

\bibitem{zhu2018bidirectional}
Lei Zhu, Zijun Deng, Xiaowei Hu, Chi-Wing Fu, Xuemiao Xu, Jing Qin, and Pheng-Ann Heng.
\newblock Bidirectional feature pyramid network with recurrent attention residual modules for shadow detection.
\newblock In {\em Proceedings of the European Conference on Computer Vision (ECCV)}, pages 121--136, 2018.

\end{thebibliography}
}

\end{document}